\pdfoutput=1

\documentclass[11pt]{article}

\usepackage{emnlp2021}

\usepackage{times}
\usepackage{latexsym}
\usepackage{caption}
\usepackage{helvet} 
\usepackage{courier}  
\usepackage{graphicx, subfig}
\usepackage{amsmath}
\usepackage{mathrsfs}
\usepackage{multirow}
\usepackage{xcolor}

\usepackage{booktabs}

\usepackage[T1]{fontenc}

\usepackage[utf8]{inputenc}

\usepackage{microtype}

\title{MapRE: An Effective Semantic Mapping Approach for Low-resource Relation Extraction}

\author{Manqing Dong, Chunguang Pan, and Zhipeng Luo \\
  DeepBlue Technology (Shanghai) Co., Ltd \\
  \texttt{dongmanqing@gmail.com, luozp@deepblueai.com} \\}

\begin{document}
\maketitle
\begin{abstract}
Neural relation extraction models have shown promising results in recent years; however, the model performance drops dramatically given only a few training samples. 
Recent works try leveraging the advance in few-shot learning to solve the low resource problem, where they train label-agnostic models to directly compare the semantic similarities among context sentences in the embedding space. 
However, the label-aware information, i.e., the relation label that contains the semantic knowledge of the relation itself, is often neglected for prediction.
In this work, we propose a framework considering both label-agnostic and label-aware semantic mapping information for low resource relation extraction.
We show that incorporating the above two types of mapping information in both pretraining and fine-tuning can significantly improve the model performance on low-resource relation extraction tasks. 
\end{abstract}

\section{Introduction}
Relation Extraction (RE), which aims at discovering the correct relation between two entities in a given sentence, is a fundamental task in NLP~\cite{gao2019fewrel}.
The problem is generally regarded as a supervised classification problem by training on large-scale labelled data~\cite{zhang2017position}. Neural models, e.g. RNN-based methods~\cite{zhou2016attention}, or more recently, BERT-based methods~\cite{soares2019matching,peng2020learning}, have shown promising results on RE tasks, where they achieve state-of-the-art performance or even comparable with human performance on several public RE benchmarks.

\begin{figure}[t]
    \centering
    \includegraphics[width=0.98\linewidth]{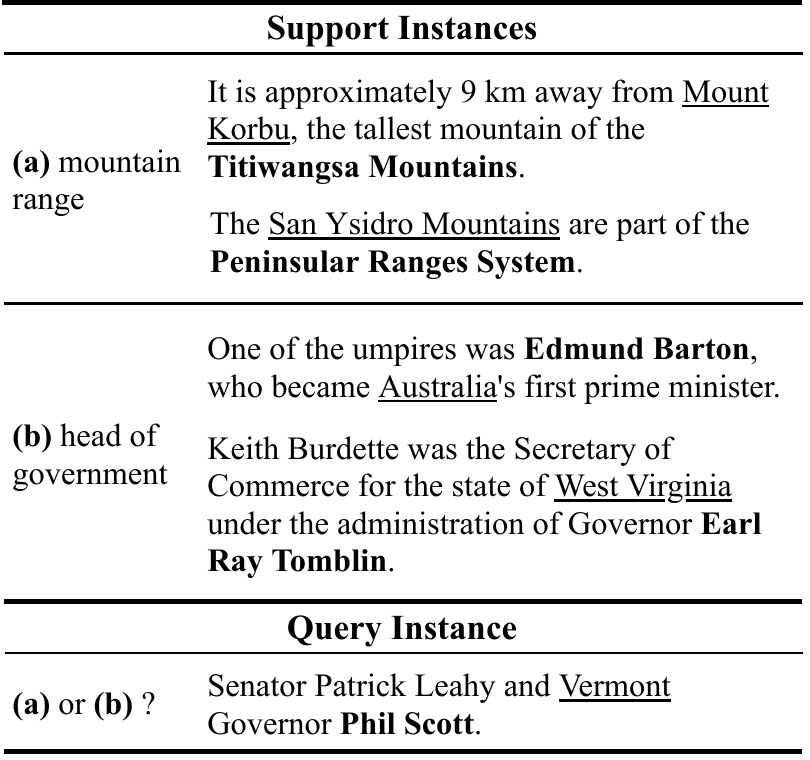}
    \caption{Example for a 2-way 2-shot relation extraction task. The entities with \underline{underlines} are head entities, and the entities in \textbf{bold} are tail entities. The target is to predict the relation between the head and the tail entities for a given query instance.}
    \label{fig:eg_fewshot}
\end{figure}

Despite the promising performance of the existing neural relation classification frameworks, recent studies~\cite{han2018fewrel} found that the model performance drops dramatically as the number of instances for a relation decreases, e.g., for long-tail relations.
An extreme condition is few-shot relation extraction, where only few support examples are given for the unseen relations, see Figure~\ref{fig:eg_fewshot} as an example.

\begin{figure*}[t]
    \centering
    \includegraphics[width=\textwidth]{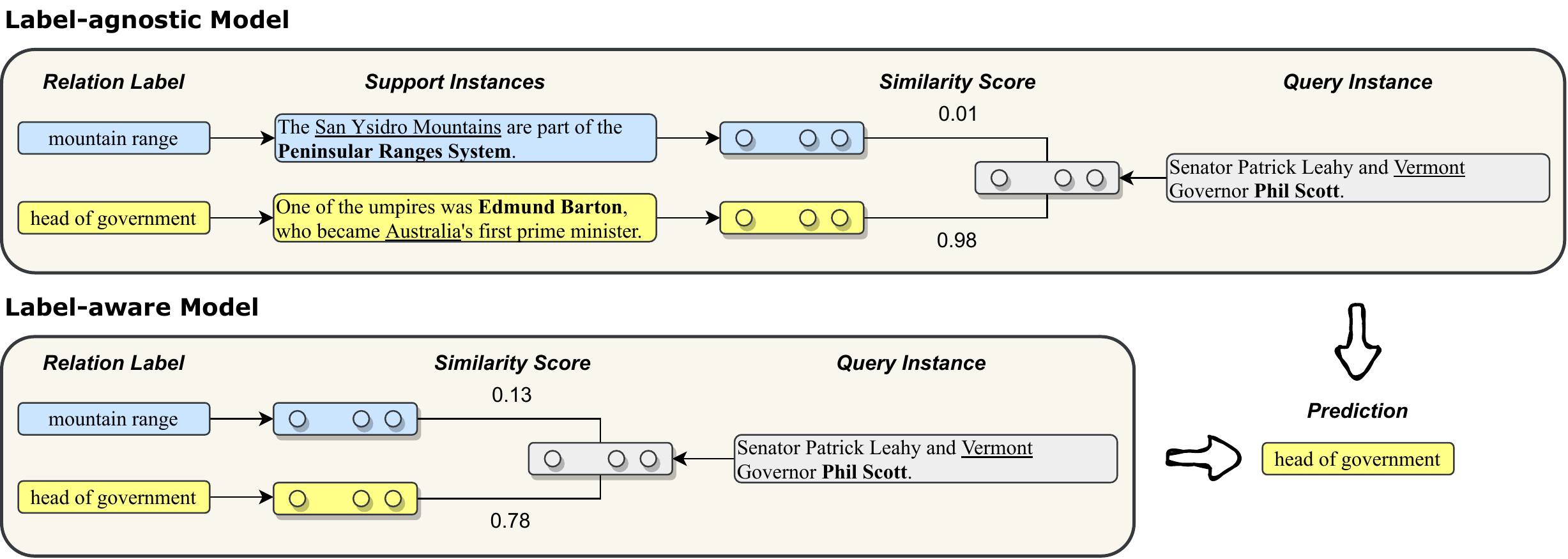}
    \caption{Examples for label-agnostic and label-aware models to relation extraction.}
    \label{fig:example_past}
\end{figure*}

A conventional way to solve the data deficiency problem of RE is distant supervision~\cite{mintz2009distant,hu2019improving}, which assumes same entity-pairs have same relations in all sentences so that to augment training data for each relation from external corpus.
However, such an approach can be rough and noisy since same entity-pairs may have different relations given different contexts~\cite{ye2019multi,peng2020learning}. 
Besides, distant supervision may exacerbate the long-tail problem in RE for the relations with only a few instances.

Inspired by the advances in few-shot learning~\cite{nichol2018first,mishra2018simple}, 
recent attempts adopt metric-based meta-learning frameworks~\cite{snell2017prototypical,koch2015siamese} to few-shot RE tasks~\cite{gao2019fewrel,ye2019multi}.
The key idea is to learn a label-agnostic model 
that compares the similarity between the query and support samples in the embedding space (see Figure~\ref{fig:example_past} for an example).
In this way, the target for RE changes from learning a general and accurate relation classifier to learning a projection network that maps the instances with the same relation into close regions in the embedding space.

Recent metric-based relation extraction frameworks~\cite{peng2020learning,soares2019matching} achieve the state-of-the-art on low-resource RE benchmarks.
However, 
these approaches are not applicable when there is no support instance for the unseen relations, since they need at least one support example to provide the similarity score of a given query sentence. 
Besides, most of the existing few-shot RE frameworks neglect the relation label for prediction, whereas the relation label contains valuable information that implies the semantic knowledge between the two entities in a given sentence.
In this work, we propose a semantic mapping framework, MapRE, which leverages both label-agnostic and label-aware knowledge. Specifically, we hope two types of matching information, i.e., the context sentences and their corresponding relation 
label (label-aware) as well as the context sentences denoting the same relations (label-agnostic), to be close in the embedding space. We show that leveraging the label-agnostic and label-aware knowledge in pretraining improves the model performance in low-resource RE tasks, and utilizing the two types of information in fine-tuning can further enhance the prediction results.
With the contribution of the label-agnostic and label-aware information in both pretraining and fine-tuning, we achieve the state-of-the-art in nearly all settings of the low-resource RE tasks (e.g., we improve the SOTA on two 10-way 1-shot datasets by 1.98\% and 2.35\%, respectively). 

Section~\ref{sec:related_work} summarizes the related work and briefly introduces the difference between our proposed method and the others. Section~\ref{sec:pretraining} illustrates the pretraining framework with considering both label-agnostic and label-aware information. We evaluate the proposed model on supervised RE in Section~\ref{sec:supervised} and few \& zero-shot RE in Section~\ref{sec:few_and_zero}, and leave concluding remarks in Section~\ref{sec:conclusion}. 

\section{Related Work}
\label{sec:related_work}
\paragraph{Meta-learning} 
One branch of meta-learning is optimization-based frameworks~\cite{nichol2018first}, e.g. model-agnostic meta-learning (MAML) ~\cite{finn2017model}, which learn a shared parameter initialization across training tasks to initialize the model parameters of testing tasks. However, a single shared parameter initialization cannot fit diverse 
task distribution~\cite{hospedales2020meta}; besides, the gradient updating strategies for the sharing parameters are complex and will take more computation resources.
Metric-based meta-learning approaches~\cite{snell2017prototypical,koch2015siamese} learn a projection network that maps the support and the query samples into the same semantic space to compare the similarities.
The metric-based approaches are non-parametric, easier for implementation, and less computationally expensive; they have shown better performance than the optimization-based approaches on a series of few-shot learning tasks~\cite{triantafillou2019meta},
thus have been widely used in recent few-shot RE frameworks~\cite{ye2019multi}. 

\paragraph{Few-shot RE} 
Prototypical network~\cite{snell2017prototypical} is probably the most widely used metric-based meta-learning framework for few-shot RE. It learns a prototype vector for each relation with a few examples, then compares the similarity between the query instance and the prototype vectors of the candidate relations for prediction~\cite{han2018fewrel}.
For example, \citet{gao2019fewrel} proposed hybrid attention-based prototypical networks to handle noisy training samples in few-shot learning. \citet{ye2019multi} further propose a multi-level matching and aggregation network for few-shot RE. 
Recent studies~\cite{soares2019matching,peng2020learning} also suggest the effectiveness of applying the metric-based approaches on pretrained models~\cite{devlin2019bert}, where optimizing the matching information between the support and query instances in embedding space obtained from the pretrained models can improve the model performance on the few-shot RE tasks.
However, the metric-based approaches are not applicable for zero-shot learning scenarios, since they need at least one example for each support instance. To fill in this gap, we propose a semantic mapping framework that leverages both label-aware and label-agnostic information for relation extraction.

\paragraph{Zero-shot learning}
An extreme condition of few-shot learning is zero-shot learning, where there is no instance provided for the candidate labels.
A standard approach is to match the inputs with the predefined label vectors~\cite{xian2017zero,rios2018few,xie2019attentive}, which assumes the label vectors take an equally crucial role as the representations of the support instances~\cite{yin2019benchmarking}. 
The label vectors are often obtained by pretrained word embeddings such as GloVe embeddings~\cite{pennington2014glove} and will be directly used for prediction~\cite{rios2018few,wang2018joint}.
For example, \citet{xia2018zero} study the zero-shot intent detection problem: they use the sum of the word embeddings as the representation for each intent label, and the prediction is based on the similarity between the inputs and the intent representations. 
\citet{zhang2019integrating} enrich the label representation with external knowledge such as the label description and the label hierarchy.
However, the label representations are fixed in most existing zero-shot learning approaches, which will lead the input-representation-learning model overfit to the label representations. 
Besides, the superiority of the label-aware models are somewhat limited to zero-shot learning scenarios -- according to our experimental results on FewRel dataset~\cite{han2018fewrel} (refer to Table~\ref{tab:variants}), the label-agnostic models perform better than the label-aware models once given support examples. 
To overcome the above issues, we propose a pretraining framework considering both label-aware and label-agnostic information for low-resource RE tasks, where the label representations are obtained via a learnable BERT-based~\cite{devlin2019bert} model.

\paragraph{RE with external knowledge}
Some works try leveraging external knowledge to address the low-resource RE tasks. For example, \citet{cetoli2020exploring} formalize RE as a question-answering task: they fine-tune on a BERT-based model that pretrained on SQUAD~\cite{rajpurkar2016squad} then use the BERT-based model to generate the prediction for the relation label.
\citet{qu2020few} follows the key idea of zero-shot learning by introducing knowledge graphs to obtain the relation label representations.
Both works show good performance on low-resource RE tasks while need extra knowledge to fine-tune the framework. However, the extra knowledge is not always available for all cases. In this work, we focus on enhancing the generalization ability of the model without referring to external knowledge, where we obtain SOTA performance on most low-resource RE benchmarks.

\color{black}
\section{Pretraining with Semantic Mapping}
\label{sec:pretraining}
\subsection{Preliminary}
\paragraph{Task definition}
Each instance $x=(c, p_h, p_t)$ includes a triple of context sentence tokens $c=[c_{0} \dots c_{m}]$ and the head and tail entity positions, where $c_{0}=\texttt{[CLS]}$ and $c_{m}=\texttt{[SEP]}$ are two special tokens denoting the start and the end of the sequence, $p_h= (p^s_h, p^e_h)$ and $p_t = (p^s_t, p^e_t)$ are the indices for head and tail entities with $ 0 \leq p^s_h \leq p^e_h \leq m$ and $0 \leq p^s_t \leq p^e_t \leq m$. 
For a \textbf{supervised learning} problem, given $\mathcal{N}$ relations $\mathcal{R}=\{r_1, \dots, r_{\mathcal{N}}\}$ and the instances for each relation, our target is to predict the correct relations $r\in\mathcal{R}$ for the testing instances. 
For a \textbf{$N$-way $K$-shot learning} problem, given support instances $\mathcal{S}=\{x_{r}^j|r \in R, j=\{1,\dots,K\}\}$ with $N$ relations $R=\{r_{1},\dots,r_{N}\}$ and $K$ examples for each relation, our target is to predict the correct relation $r\in R$ of the entities for a query instance $x_q$.

\begin{figure*}[t]
    \centering
    \includegraphics[width=\textwidth]{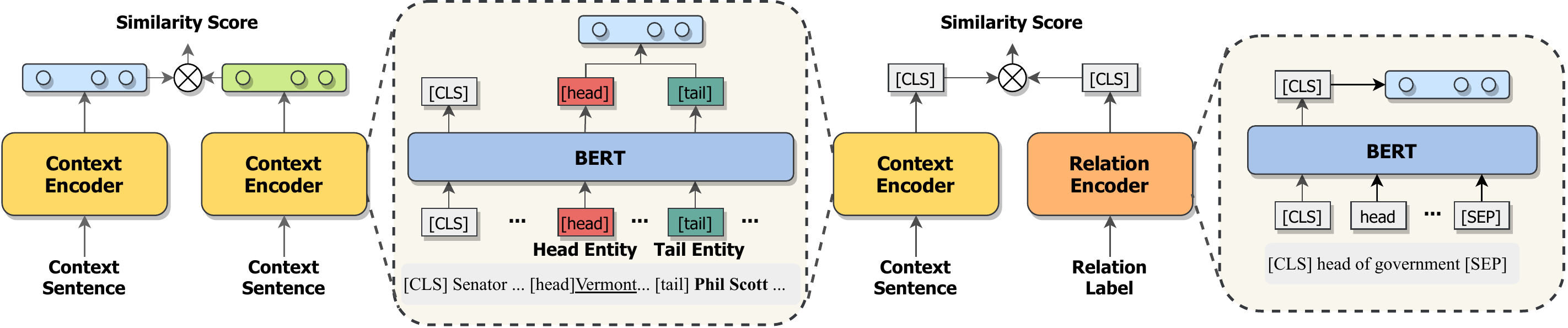}
    \caption{The pretraining framework for MapRE, where we consider both label-agnostic and label-aware semantic mapping information in training the whole framework. }
    \label{fig:pretrain_example}
\end{figure*}

\paragraph{Differences between supervised RE and few-shot RE}
There are several differences between supervised RE and few-shot RE. First, supervised RE tries to learn a $\mathcal{N}$-way relation classifier that could fit all training instances, while few-shot RE tries to learn a $N$-way classifier (normally $N \ll \mathcal{N}$) by learning from only a few samples. Second, the training and testing data for few-shot RE have no intersection in relation types, i.e. during the testing phase, the model is required to generalize to unseen labels with only a few samples.

\paragraph{Pretraining for low-resource RE}
Recent studies~\cite{soares2019matching,peng2020learning} find that pretrain the model with contrastive ranking loss~\cite{sohn2016improved,oord2018representation} can improve the generalization ability of the model in low-resource RE tasks. The key idea is reducing the semantic gap between the instances with the same relations in the embedding space.
In other words, instances with same relations should have similar representations.

\subsection{Matching Sample Formulation}
Following the idea of~\citet{soares2019matching} and~\citet{peng2020learning}, we construct mapping functions for relation extraction. 
Specially, we hope two types of matching samples to be close in the semantic space: \textbf{1)} the context sentences denoting same relations, and \textbf{2)} the context sentences and the corresponding relation labels.

Given a knowledge graph $\mathcal{G}$ containing extensive examples of relation triples $T=(h, r, t), T\in \mathcal{G}$, we will first randomly sample the relation triples; then, sentences containing the same head $h$ and tail $h$ entities and denoting the same relation $r$ will be sampled from the corpus for this triple, i.e. $\{x=(c,p_h,p_t)|x\in T\}$.
Specially, at each sampling step, $N$ triples with $N$ different relations $\{r_i|i=1,\dots,N\}$ are sampled from $\mathcal{G}$. For each triple $T=(h,r,t)$, a pair of sentences $\{(x_A,x_B)|x_A,x_B\in T\}$ will be extracted from the corpus, so that we have $2N$ sentences in total. 
For each sentence, we take a similar strategy as in \cite{soares2019matching,peng2020learning} that a probability of 0.7 is set to mask the entity mentions when fed into the sentence context encoder to avoid the model memorizes the entity mentions or shallow cues during pretraining.

Suppose the sentence context encoder is denoted as $f_\texttt{CON}$, and the relation encoder is denoted as $f_\texttt{REL}$, we hope the semantic gap between each pair of sentences that denote for same relation, i.e., $d(f_\texttt{CON}(x_A), f_\texttt{CON}(x_B)), x_A,x_B\in T$, and the semantic gap between the context sentences and their relation labels, i.e., $d(f_\texttt{CON}(x_A),f_\texttt{REL}(r))$ and $ d(f_\texttt{CON}(x_B),f_\texttt{REL}(r))$, to be small in in the embedding space.
Figure~\ref{fig:pretrain_example} shows an example of the matching samples, where both the context encoder $f_\texttt{CON}$ and the relation encoder $f_\texttt{REL}$ are a BERT$_\texttt{BASE}$ model~\cite{devlin2019bert}. 
According to~\citet{soares2019matching}, the concatenation of the special tokens (i.e., \texttt{[head]} and \texttt{[tail]}) at the start of the head and the tail entities, provides best performance for downstream relation classification tasks, thus we take $f_\texttt{CON}(x)[\texttt{[head]},\texttt{[tail]}]$ to compare the label-agnostic similarities between sentences. 
We use the embedding of the special $\texttt{[CLS]}$ token in the context encoder $f_{\texttt{CON}}(x)[\texttt{CLS}]$ to denote the label-aware information for the context sentence, and the $\texttt{[CLS]}$ token in relation encoder $f_{\texttt{REL}}(r)[\texttt{CLS}]$ to denote the relation representation. This is to avoid the override of the memorization in the head and tail special tokens and to improve the generalization ability of the sentence context encoder. 
Another reason is the dimension of the concatenation  $[[\texttt{head}],[\texttt{tail}]]$ and the $[\texttt{CLS}]$ token does not match, which needs extra parameter space to optimize. The extra parameter space can be easily over-fitted to training data and produce biased prediction performance when distinct distribution between the training and testing sets exists.

\subsection{Training Objectives}
\begin{figure}[ht]
    \centering
    \includegraphics[width=\linewidth]{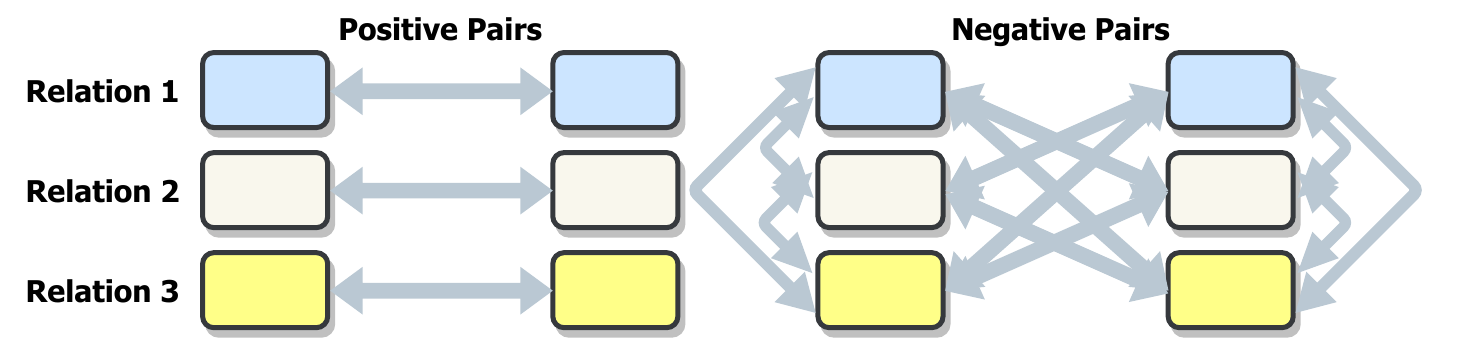}
    \caption{Examples for positive and negative sentence context representation pairs.}
    \label{fig:contrastive_pairs}
\end{figure}

At each sampling step, we have $2N$ sentences with $N$ pairs of sentences denoting $N$ distinct relations. 
For each sentence $x$, we get its context embedding $\textbf{u} = f_{\texttt{CON}}(x)[\texttt{[head]},\texttt{[tail]}]$ and its label-aware embedding $\textbf{w}=f_{\texttt{CON}}(x)[\texttt{CLS}]$. The corresponding relation representation is obtained by $\textbf{v}=f_{\texttt{REL}}(r)[\texttt{CLS}]$. 
We use contrastive training~\cite{oord2018representation,chen2020simple} to train the MapRE, which pulls the 'neighbors' together and pushes 'non-neighbors' apart. 
Specifically, we consider three training objectives to optimize the whole framework.
\paragraph{Contrastive Context Representation Loss}
We follow the work by~\cite{peng2020learning} to calculate the contrastive loss of the sentence context representations~\footnote{\url{https://kevinmusgrave.github.io/pytorch-metric-learning/losses/\#ntxentloss}}.
For example, for sentence $x_A^i$ from the positive pair $(x^i_A, x^i_B)$ (both represents relation $r_{i}$), any sentence in other pairs forms the negative pair with $x_A^i$, i.e., $(x^i_A,x^j_B)$ and $(x_A^i,x^j_A)$, for $1\leq j\leq N, j\neq i$ (examples are shown in Figure~\ref{fig:contrastive_pairs}).
Then for $x_A^i$, we maximize
$\frac{\exp{({\textbf{u}^{i}_A}^\top \textbf{u}_B)}}{\Sigma_{j}\exp{({\textbf{u}^{i}_A}^\top \textbf{u}^j_B)} + \Sigma_{j}\exp{({\textbf{u}^{i}_A}^\top \textbf{u}^j_A})}$.
Sum the log loss for each sentence, we get the contrastive context representation loss as $\mathcal{L}_{CCR}$.

\paragraph{Contrastive Relation Representation Loss}
We also calculate the contrastive loss between the label-aware representation $\textbf{w}$ and the relation representations $\textbf{v}$. For the $2N$ sampled sentences of $N$ relations, we hope to minimize the loss
\begin{equation}
\mathcal{L}_{CRR} = -\Sigma^{2N}_{i=1} \log\frac{\exp{(\textbf{w}_{i}^{\top}\textbf{v}_i)}}{\Sigma^{N}_{j=1} \exp{(\textbf{w}_{i}^{\top}\textbf{v}_j)}}.
\end{equation}

\paragraph{Masked Language Modeling (MLM)}
We also consider the conventional Masked Language Modeling objective~\cite{devlin2019bert}, which randomly masks tokens in the inputs and predicts them in the outputs to let the context encoder engaging more semantic and syntactic knowledge. Denoting the loss by $\mathcal{L}_{MLM}$, the overall training objective is
\begin{equation}
    \mathcal{L} = \mathcal{L}_{CCR} + \mathcal{L}_{CRR} + \mathcal{L}_{MLM}
\end{equation}

We pretrain the whole framework on Wikidata~\cite{vrandevcic2014wikidata} with a similar strategy as in~\cite{peng2020learning}, where we exclude any overlapping data between Wikidata and the datasets for further experiments. 

\section{Supervised RE}
\label{sec:supervised}
\subsection{Fine-tuning for supervised RE}
\label{sec:fine-tune_supervised}
We obtain a pretrained context encoder $f_{\texttt{CON}}$ and a relation encoder $f_{\texttt{REL}}$ after the pretraining process mentioned above.
A conventional way for supervised RE is to append several fully connected layers to the context encoder $f_{\texttt{CON}}$ for classification, which can also be regarded as computing the similarity between the output of the context encoder and the one-hot relation label vectors (see the left part of Figure~\ref{fig:structure_supervised} as an example).
Instead of using one-hot representation for the relation labels, we use the relation representation obtained from the relation encoder to calculate the similarities. 
An example is shown in the right part of Figure~\ref{fig:structure_supervised}. 
The prediction is made by
\begin{equation}
    \hat{r} = \mathop{\arg\max}_{r} \frac{\exp{(\sigma(f_{\texttt{CON}}(x))^{\top}f_{\texttt{REL}}(r))}}{\Sigma_{r'\in \mathcal{R}}\exp{(\sigma(f_{\texttt{CON}}(x))^{\top}f_{\texttt{REL}}(r'))}}
\end{equation}
where $\sigma$ stands for fully connected layers, $f_{\texttt{REL}}(r)$ denotes the embedding of the special token \texttt{[CLS]} in the relation encoder, and $f_{\texttt{CON}}(x)$ here outputs the concatenation of the special tokens of head and tail entities $[\texttt{[head]}, \texttt{[tail]}]$.
We optimize the context encoder, relation encoder, and the fully connected layers with cross-entropy loss for supervised training.

\subsection{Evaluation}
\paragraph{Datasets}
We evaluate on two benchmark datasets, ChemProt~\cite{kringelum2016chemprot} and Wiki80~\cite{han2019opennre} for supervised RE tasks. The former includes 56,000 instances for 80 relations, and the latter includes 10,065 instances for 13 relations.

\begin{figure}
    \centering
    \includegraphics[width=\linewidth]{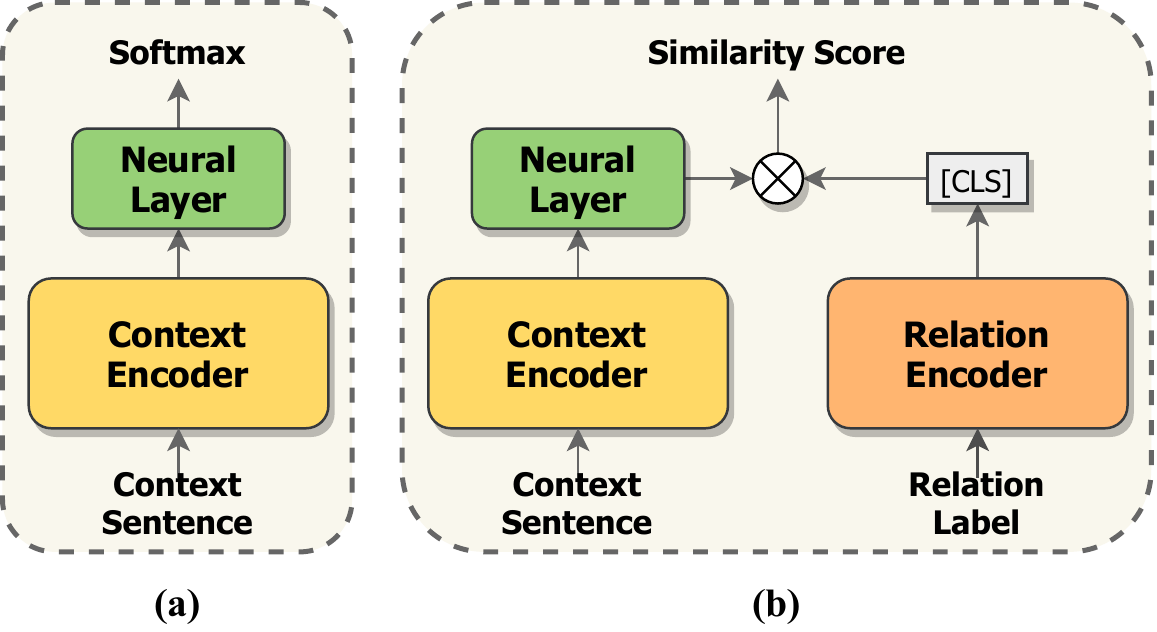}
    \caption{The frameworks for supervised learning. \textbf{Left}: uses fully connected layers to predict the probability distribution over all relations, used in BERT, MTB, CP, and MapRE-L. \textbf{Right}: compares the sentence context embedding with the relation representations, and regards the relation with highest similarity score as the prediction, used in MapRE-R.}
    \label{fig:structure_supervised}
    \vspace{-0.2cm}
\end{figure}

\paragraph{Comparison Methods}
Numerous studies have been done for supervised RE tasks. Here we focus on low-resource RE and choose the following three representative models for comparison. 1) BERT~\cite{devlin2019bert}: the widely used pretrained model for NLP tasks. In this case, the model takes the embedding of the special tokens of the head and tail entities for prediction via several fully connected layers, similar to the conventional strategy shown in the left part of the Figure~\ref{fig:structure_supervised}. 
2) MTB~\cite{soares2019matching}: a pretrained framework for RE, which regards the sentences with the same head and tail entities as positive pairs. The fine-tuning strategy is same as in BERT. 
3) CP~\cite{peng2020learning}: a pretrained framework that is analogous to MTB. The difference is that the model treats sentences with the same relations as positive pairs during the pretraining phase. The fine-tuning strategy is the same as BERT and MTB.

\begin{table}
    \centering
    \small
    \begin{tabular}{c|c|ccc}
    \toprule
    \textbf{Dataset} & \textbf{Method} & \textbf{1\%} & \textbf{10\%} & \textbf{100\%} \\ \midrule
    \multirow{4}{5em}{\textbf{Wiki80}} & BERT & 0.559 & 0.829 & 0.913 \\
    & MTB & 0.585 & 0.859 & 0.916 \\
    & CP & 0.827 & 0.893 & 0.922 \\ 
    & \textbf{MapRE-L} & \textbf{0.850} & \textbf{0.915} & \textbf{0.933} \\
    & \textbf{MapRE-R} & \textbf{0.904} & \textbf{0.921} & \textbf{0.933} \\ \midrule
    \multirow{4}{5em}{\textbf{ChemProt}} & BERT & 0.362 & 0.634 & 0.792 \\
    & MTB & 0.362 & 0.682 & 0.796 \\
    & CP & 0.361 & \textbf{0.708} & 0.806 \\ 
    & \textbf{MapRE-L} & \textbf{0.424} & 0.666 & \textbf{0.813} \\
    & \textbf{MapRE-R} & \textbf{0.416} & 0.693 & \textbf{0.814} \\ \bottomrule
    \end{tabular}
    \caption{Comparison results on supervised learning tasks in accuracy. 1\%, 10\%, and 100\% denote the proportion of the training data used for fine-tuning.}
    \label{tab:supervised_results}
    \vspace{-0.4cm}
\end{table}

\paragraph{Comparison Results}
Table~\ref{tab:supervised_results} shows the comparison results on the two datasets with training on different proportions of the training sets. For our model, we consider the model performance with different fine-tuning strategies as shown in the left and right part in Figure~\ref{fig:structure_supervised}. We denote the two variants as MapRE-L and MapRE-R. The detailed parameter settings can be found in the Appendix. 
We can observe that: 1) pretraining on the BERT with matching information (i.e., MTB, CP, and our MapRE) can improve the model performance on low-resource RE tasks; 2) comparing MapRE-L with CP and MTB, adding the label-aware information during pretraining can significantly improve the model performance, especially on extremely low-resource conditions, e.g., when only 1\% of training sets are available for fine-tuning; and 3) comparing MapRE-R with MapRE-L, which also considers the label-aware information in fine-tuning, shows better and more stable performance in most conditions. Overall, the results suggest the importance of engaging the label-aware information in pretraining and fine-tuning to improve the model performance on low-resource supervised RE tasks.

\section{Few \& Zero-shot RE}
\label{sec:few_and_zero}
\subsection{Fine-tuning for few-shot RE}
\label{sec:finetune_fewshot}
\begin{figure}[t]
    \centering
    \includegraphics[width=\linewidth]{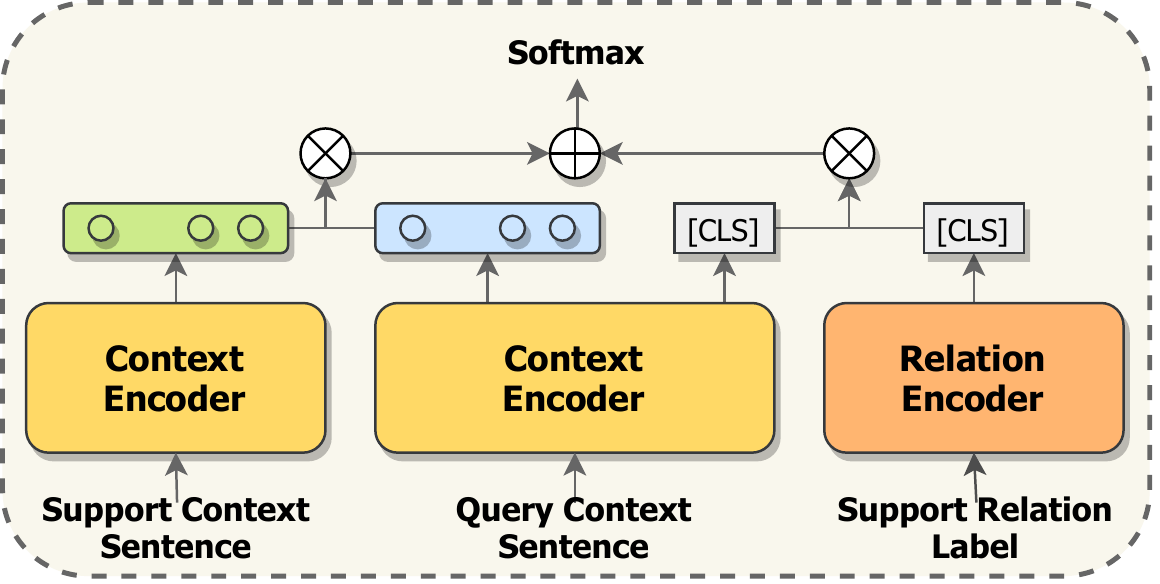}
    \caption{The framework for few-shot learning with MapRE. Both label-agnostic information, i.e., the matching information among the context sentence representations, and label-aware information, i.e., the semantic gap between the sentence label-aware representation and the relation label representation, are considered for fine-tuning.}
    \label{fig:structure_fewshot}
\end{figure}

In the case of few-shot learning, the model is required to predict for new instances with only a few given samples. For a $N$-way $K$-shot problem, the support set $\mathcal{S}$ contains $N$ relations that each is with $K$ examples, and the query set contains $Q$ samples that each belongs to one of the $N$ relations.
To fine-tune the model for few-shot RE, we construct the training set in a series of $N$-way $K$-shot learning tasks. For each task, the prediction for a query instance $x_q$ is made by comparing the \textbf{label-agnostic mapping information}, i.e., the similarity between the query context sentence representation $u_q$ and the support context sentence representation $u_r$, as well as the \textbf{label-aware mapping information}, i.e., the semantic gap between the query label-aware representation $w_q=f_{\texttt{CON}}(x_q)[\texttt{CLS}]$ and the relation label representation $v_r=f_{\texttt{REL}}(r)[\texttt{CLS}]$:
\begin{align}
    \label{eq:alpha_beta} \hat{r} & = \mathop{\arg\max}_{r} \frac{\exp(\alpha \cdot u^{\top}_q u_{r} + \beta \cdot w^{\top}_q v_r)}{\Sigma_{r'\in \mathcal{R}}\exp(\alpha \cdot u^{\top}_q u_{r'} + \beta \cdot w^{\top}_q v_{r'})} \\
    u_r & = \frac{1}{K}\Sigma_{k} u^k_r = \frac{1}{K}\Sigma_{k} f_{\texttt{CON}}(x^k_{r})[[\texttt{head}],[\texttt{tail}]] 
\end{align}
where $u_r$ is the prototype sentence representation for $K$ support instances denoting relation $r$; $\alpha$ and $\beta$ are two learnable coefficients controlling the contribution of the two types of semantic mapping information. An example of the few-shot learning framework is shown in Figure~\ref{fig:structure_fewshot}.
We update both context encoder and relation encoder with cross-entropy loss on the generated $N$-way $K$-shot training tasks. We use dot product as the measurement of the similarities, which shows the best performance compared with other measurements. Details about the model settings can be found in the Appendix.  

\subsection{Evaluation}

\begin{table*}
    \centering
    \small
    \begin{tabular}{c|cccc|cccc}
    \toprule
        \multirow{3}{4em}{\textbf{Method}} & \multicolumn{4}{c|}{\textbf{FewRel}} & \multicolumn{4}{c}{\textbf{NYT-25}} \\ \cline{2-9}
        & 5-way  & 5-way  & 10-way & 10-way  & 5-way  & 5-way  & 10-way & 10-way  \\ 
        & 1-shot & 5-shot & 1-shot & 5-shot  & 1-shot & 5-shot & 1-shot & 5-shot \\ \hline
        Proto    & 80.68 & 89.60 & 71.48 & 82.89 & 77.63 & 87.25 & 66.49 & 79.51 \\
        BERT-pair& 88.32 & 93.22 & 80.63 & 87.02 & 80.78 & 88.13 & 72.65 & 79.68 \\
        REGRAB   & 90.30 & 94.25 & 84.09 & 89.93 & 89.76 & 95.66 & 84.11 & 92.48 \\
        MTB      & 91.10 & 95.40 & 84.30 & 91.80 & 88.90 & 95.53 & 83.08 & 92.23 \\ 
        CP       & 95.10 & 97.10 & 91.20 & 94.70 & 91.08 & 94.73 & 83.99 & 90.18 \\ 
         \hline
        \textbf{MapRE} & \textbf{95.73} & \textbf{97.84} & \textbf{93.18} & \textbf{95.64} & \textbf{91.90} & \textbf{96.01} & \textbf{86.46} & \textbf{92.68} \\
        \bottomrule
    \end{tabular}
    \caption{Comparison results on the test set of the FewRel and NYT-25 datasets in accuracy. }
    \label{tab:results_fewshot}
\end{table*}

\paragraph{Datasets}
We evaluate the proposed method on two few-shot learning benchmarks: FewRel~\cite{han2018fewrel} and NYT-25~\cite{gao2019fewrel}. The FewRel dataset consists of 70,000 sentences for 100 relations (each with 700 sentences) derived from Wikipedia. There are 64 relations for training, 16 relations for validation, and 20 relations for testing. The testing dataset contains 10,000 query sentences that each is given $N$-way $K$-shot relation examples and has to be evaluated online (the labels for the testing set is not published). The NYT-25 dataset is a processed dataset by ~\cite{gao2019fewrel} for few-shot learning. We follow the preprocessing strategy by~\citet{qu2020few} to randomly sample 10 relations for training, 5 for validating, and 10 for testing.

\paragraph{Comparison methods}
Many recent studies try employing the advances of meta-learning~\cite{hospedales2020meta} to few-shot RE tasks. We consider the following representative methods for comparison.
1) \textbf{Proto}~\cite{han2018fewrel} is a work using Prototypical Networks~\cite{snell2017prototypical} for few-shot RE. The model tries to find the prototypical vectors for each relation  from supporting instances, and compares the distance between the query instance and each prototypical vector under certain distance metrics. Each instance is encoded by a BERT$_{\texttt{BASE}}$ model. 
2) \textbf{BERT-pair}~\cite{gao2019fewrel} is a BERT-based model that encodes a pair of sentences to a probability that the pair of sentences expressing the same relation. 
3) \textbf{REGRAB}~\cite{qu2020few} is a label-aware approach that predicts the relations based on the similarity between the context sentence and the relation label. The relation label representation is initialized via an external knowledge graph, where a Bayesian meta-learning approach is further used to infer the posterior distribution of the relation representation. The representation of the context sentence is learned by a BERT$_{\texttt{BASE}}$ model.
4) \textbf{MTB}~\cite{soares2019matching} is a pretraining framework with the assumption that the sentences with the same head and tail entities are positive pairs. During the testing phase, it ranks the similarity score between the query instance and the support instances and chooses the relation with the highest score as the prediction.
5) \textbf{CP}~\cite{peng2020learning} is also a pretraining framework that regards the sentences with the same relations as positive pairs. The fine-tuning strategy of CP is much like the strategy in Proto; the difference is that they use the dot product instead of Euclidean distance to measure the similarities between instances. Our method differs from CP in that we also consider label-aware information in both pretraining and fine-tuning.

\paragraph{Comparison results}
We consider four types of few-shot learning tasks in our experiments, which are 5-way 1-shot, 5-way 5-shot, 10-way 1-shot, and 10-way 5-shot learning tasks. For the comparison methods, most results are collected from the published papers~\cite{gao2019fewrel,peng2020learning,qu2020few}. While for MTB~\cite{soares2019matching}, which does not have publicly available code for reproduction, we present the results reproduced with a BERT$_{\texttt{BASE}}$ model trained with the MTB pretraining strategies~\cite{soares2019matching,peng2020learning}. As for CP~\cite{peng2020learning}, which does not include the results for the NYT-25 dataset, we reproduce the results by fine-tuning the pretrained CP~\footnote{\url{https://github.com/thunlp/RE-Context-or-Names}} on the NYT-25 datasets.
For our model, we fine-tune on our pretrained MapRE with the approaches described in Section~\ref{sec:finetune_fewshot}, which considers both label-agnostic and label-aware information in fine-tuning. More details about the parameter settings can be found in the Appendix. 
Table~\ref{tab:results_fewshot} presents the comparison results on two few-shot learning datasets in different task settings. We can observe that, pretraining the framework with matching information between the instances (i.e., MTB, CP, and ours) can significantly improve the model performance in few-shot scenarios. 
Comparing the label-aware methods (i.e., REGRAB and ours) with label-agnostic methods on the NYT-25 dataset, which lies in a different domain than Wikipedia, the label-aware methods can grasp more hints from the relation semantic knowledge for prediction. Such improvements become much significant with a larger number of relations $N$ and fewer support instances $K$, which suggests that the label-aware information is valuable in extreme low-resource conditions. For all settings, the proposed MapRE, which considers both label-agnostic and label-aware information in pretraining and fine-tuning, provides steady 
performance and outperforms a series of baseline methods as well as the state-of-the-art. The results prove the effectiveness of the proposed framework, and suggest the importance of the semantic mapping information from both label-aware and label-agnostic knowledge. 

\paragraph{Discussion}
We further consider two variants of MapRE, i.e., employing only the label-agnostic information or only the label-aware information, to discover how the two types of information contribute to the final performance. 
\begin{table}[h]
    \centering
    \small
    \begin{tabular}{c|cccc}
    \toprule
        \multirow{3}{4em}{\textbf{Method}} & \multicolumn{4}{c}{\textbf{FewRel}} \\ \cline{2-5}
        & 5-way  & 5-way  & 10-way & 10-way  \\ 
        & 1-shot & 5-shot & 1-shot & 5-shot  \\ \hline
        Label-agnostic & 95.56 & 97.60 & 92.55 & 95.19 \\
        Label-aware & 72.97 & 72.74 & 61.05 & 60.98 \\ 
        Both & 95.73 & 97.84 & 93.18 & 95.64 \\
        \bottomrule
    \end{tabular}
    \caption{Accuracy on the test set of the FewRel dataset.}
    \label{tab:variants}
\end{table}
Table~\ref{tab:variants} shows the model performance on different options in fine-tuning the framework. Comparing the results of label-agnostic only MapRE with the model CP in Table~\ref{tab:results_fewshot}, where the only difference is that we consider the label-aware information in pretraining the framework, we can see that the incorporating the relation label information does help the model to capture more semantic knowledge. However, if we only consider the label-aware information in fine-tuning, the performance drops since the model does not utilize any support instances, which is much like zero-shot learning. Note that there are fluctuates in 5-way 5-shot and 10-way 5-shot of the relation-aware only MapRE; this may be caused by the difference in the testing set of the FewRel for the four few-shot learning tasks provided online~\footnote{\url{https://competitions.codalab.org/competitions/27980}}. We will discuss more details about zero-shot RE in the following subsection. The results of the label-aware only MapRE suggest the importance of the label-agnostic knowledge in few-shot RE. Overall, both label-agnostic and label-aware knowledge are valuable for few-shot RE tasks, and using them in both pretraining and fine-tuning can significantly improve the results.

\subsection{Zero-shot RE}
We further consider an extreme condition of low-resource RE, i.e., zero-shot RE, where no support instance is provided for prediction. Under the condition of zero-shot RE, most of the above few-shot RE frameworks are not applicable since they need at least one example for each support relation for comparison.
Previous studies for zero-shot learning lie in representing the label by vectors, then compare the input embedding with the label vectors for comparison~\cite{xian2017zero,rios2018few,xie2019attentive}.
\begin{table}[t]
    \centering
    \small
    \begin{tabular}{c|cc|cc}
    \toprule
    \multirow{3}{4em}{\textbf{Method}} & \multicolumn{2}{c|}{\textbf{FewRel}} & \multicolumn{2}{c}{\textbf{NYT-25}} \\ \cline{2-5}
    & 5-way  & 10-way & 5-way & 10-way \\
    & 0-shot & 0-shot & 0-shot & 0-shot \\ \hline
    \citet{qu2020few}$^{\clubsuit}$ & 52.50 & 37.50 & 40.50 & 24.50 \\
    \citet{cetoli2020exploring} & 86.00 & 76.20 & - & - \\
    \textbf{MapRE} & \textbf{90.65} & \textbf{81.46} & \textbf{72.14} & \textbf{59.94} \\ \bottomrule
    \end{tabular}
    \caption{The comparison results of the zero-shot RE on FewRel and NYT-25 datasets in accuracy. The results for the FewRel dataset and the NYT-25 dataset are evaluated on the validation set and test set, respectively.$^{\clubsuit}$ The results for \citet{qu2020few} are observed from the figures in the paper with a standard deviation of 2\%.}
    \label{tab:results_zeroshot}
\end{table}
The work by \citet{qu2020few} extends the idea by inferring the posterior of the relation label vectors initialized by an external knowledge graph. Another direction is to formalize the zero-shot RE problem as a question-answering task, where \citet{cetoli2020exploring} fine-tune on a BERT-based model pretrained on SQUAD~\cite{rajpurkar2016squad}, then use it to generate the relation prediction. Both work needs extra knowledge to tune the framework; however, the external knowledge is not always available for the given tasks. 
In our work, we fine-tune on the pretrained MapRE with only label-aware information for zero-shot learning, which can be regarded as a special case in Equation~(\ref{eq:alpha_beta}) when $\alpha=0$ and $\beta=1$. 
The results show that, compared to the two recent zero-shot RE methods, the proposed MapRE obtains outstanding performance on all zero-shot settings, which proves the effectiveness of our proposed framework.

\section{Conclusion}
\label{sec:conclusion}
In this work, we propose MapRE, a semantic mapping approach considering both label-agnostic and label-aware information for low-resource relation extraction (RE).
Extensive experiments on low-resource supervised RE, few-shot RE, and zero-shot RE tasks present the outstanding performance of the proposed framework. The results suggest the importance of both label-agnostic and label-aware information in pretraining and fine-tuning the model for low-resource RE tasks. In this work, we did not investigate the potential effect caused by the domain shift problem, and we will leave the analysis on this to future works.  

\color{black}

\bibliography{anthology,custom}
\bibliographystyle{acl_natbib}

\appendix

\section{Appendix}
\label{sec:appendix}

\subsection{Pretraining Details}
\label{sec:appendix}
\paragraph{Data preparation} We take similar strategies as in CP~\cite{peng2020learning} for pretraining the models. The difference is we also consider the label-aware information to pretrain the model.
The pretraining corpus is from Wikidata~\cite{vrandevcic2014wikidata}, where we exclude any overlapping data between Wikidata and the datasets we used for evaluation.
The training instances are sampled from the Wikidata as we described in the section of matching sample formulation. 

\paragraph{Implementation details} We train on the BERT$_\texttt{BASE}$ model from the open-source transformer toolkits~\footnote{\url{https://github.com/huggingface/transformers}} and use AdamW~\cite{loshchilov2018decoupled} as the optimizer. The max length for the input is set as 60. 
The pretraining is implemented with eight Tesla V100 32G GPUs, which will take about 6 hours for about 11,000 training steps with the first 500 steps as the warmup steps. The batch size is set as 2040, the learning rate is $3\times 10^-5$, the weight decay rate is $1\times 10^-5$, and the max gradient norm for clipping is set as 1.0. 

\subsection{Fine-tuning Details}
\paragraph{Supervised Relation Extraction}
The two supervised datasets, Wiki80 and ChemProt, can be found in the repository~\footnote{\url{https://github.com/thunlp/RE-Context-or-Names}}. We follow the same strategy to split each dataset into training, validation, and testing samples, where we have accordingly 39,200, 5,600, and 11,200 samples for Wiki80 dataset, and 4,169, 2,427, and 3,469 for the ChemProt dataset. We also follow their settings to 1\%, 10\%, and 100\% of the training sets to evaluate the model performance in low-resource scenarios.
\begin{table}[t]
    \centering
    \small
    \begin{tabular}{c|cc}
    \toprule
        \textbf{Parameter} & \textbf{Wiki80} & \textbf{ChemProt} \\ \midrule
        Batchsize & 64 & 64 \\ 
        Max training epochs & 20 & 100 \\
        Learning rate & $3\times 10^-5$ & $3\times 10^-5$ \\
        Weight decay rate & $1\times 10^-5$ & $1\times 10^-5$ \\
        Warmup steps & 500 & 500 \\
        Max sentence length & 100 & 100 \\
        \bottomrule
    \end{tabular}
    \caption{Fine-tuning settings for supervised RE.}
    \label{tab:supervised}
\vspace{-0.5cm}
\end{table}
The parameter settings to fine-tune on the two datasets can be found in table~\ref{tab:supervised}. 

\paragraph{Few \& Zero-shot Relation Extraction}
The details about the two datasets can be found in \cite{han2018fewrel,gao2019fewrel,qu2020few}. 
\begin{table}[h]
    \centering
    \small
    \begin{tabular}{c|cc}
    \toprule
        \textbf{Parameter} & \textbf{FewRel} & \textbf{NYT-25} \\ \midrule
        Training task & 5-way 1-shot & 5-way 5-shot  \\ 
        \# Training query instances & 1& 1 \\
        Max sentence length & 60 & 200 \\
        Batch size & 4 & 4 \\
        Training iteration & 10,000 & 1,000\\
        Learning rate & $3\times 10^-5$ & $3\times 10^-5$ \\
        Weight decay rate & $1\times 10^-5$ & $1\times 10^-5$ \\
    \bottomrule
    \end{tabular}
    \caption{Fine-tuning settings for few-shot RE.}
    \label{tab:fewshot}
\vspace{-0.2cm}
\end{table}
The general parameter settings for both few and zero-shot learning are shown in Table~\ref{tab:fewshot}. The difference of the settings for few and zero settings lies in the settings of the coefficients $\alpha$ and $\beta$, which controls the contribution of the relation-agnostic and relation-aware information. 
For few-shot learning, we initialize the two coefficients as 0.95 and 1.05, where they will be optimized during fine-tuning. As for the zero-shot learning, which only uses the relation-aware information, we set $\alpha$ as 0 and $\beta$ as 1.0.


\end{document}